\pdfoutput=1
\documentclass[11pt]{article}

\usepackage{acl}

\usepackage{times}
\usepackage{amsmath}
\usepackage{adjustbox}
\usepackage{latexsym}
\usepackage{numprint}
\usepackage{multirow}
\usepackage{graphicx}
\usepackage{amsfonts}
\usepackage{caption}
\usepackage{subcaption}
\usepackage{hyperref}
\usepackage[T1]{fontenc}
\usepackage[utf8]{inputenc}

\usepackage{microtype}

\title{Understanding the effects of language-specific class imbalance in multilingual fine-tuning}

\author{Vincent Jung \\
  Idiap Research Institute \\
  \texttt{vincent.jung@idiap.ch} \\\And
  Lonneke van der Plas \\
  Idiap Research Institute \\
  \texttt{lonneke.vanderplas@idiap.ch} \\}

\begin{document}
\maketitle
\begin{abstract}
We study the effect of one type of imbalance often present in real-life multilingual classification datasets: an uneven distribution of labels across languages. We show evidence that fine-tuning a transformer-based Large Language Model (LLM) on a dataset with this imbalance leads to worse performance, a more pronounced separation of languages in the latent space, and the promotion of uninformative features. We modify the traditional class weighing approach to imbalance by calculating class weights separately for each language and show that this helps mitigate those detrimental effects. These results create awareness of the negative effects of language-specific class imbalance in multilingual fine-tuning and the way in which the model learns to rely on the separation of languages to perform the task.
\end{abstract}

\section{Introduction}

Transformer-based Large Language Models (LLMs) lend themselves well to automatic classification tasks due to their superior performance, ability to be pre-trained on large amounts of data, and easy fine-tuning on downstream tasks. Recently, methods like LoRA \citep{hu_lora_2021} and Adapters \citep{houlsby_parameter-efficient_2019} have been developed to fine-tune LLMs using fewer resources, making automation of classification tasks using LLMs more accessible than ever.  Multilingual versions of large language models, such as mBERT are readily available. They are pre-trained on large multilingual corpora and build latent spaces that have both language-agnostic and language-specific components \citep{pires_how_2019}. 

Previous works have studied the effect of fine-tuning on monolingual data on the representation of the multilingual space and cross-lingual transfer performance \citep{conneau_unsupervised_2020, lampleCrosslingualLanguageModel2019a} and showed that fine-tuning on a specific task with monolingual data reduces language-specificity \citep{tanti-etal-2021-language}. What is relatively understudied is the effect of multilingual fine-tuning on the multilingual space, which is especially interesting because it is not guaranteed that labels are similarly distributed across languages which could create an incentive for the model to rely on language for predictions. Oftentimes, curated multilingual datasets will have the same distribution of labels across languages, and it is pointed out as a desirable property \cite{SCHWENK18.658}. However, in real-world datasets, data is often heterogeneous and class label distributions can vary significantly between languages. An example of this is the SemEval 2018 Task 1 dataset \citep{mohammad-etal-2018-semeval}. Class imbalance in the monolingual setting has been the focus of many previous works \citep{henning-etal-2023-survey}, some work addresses class imbalance in the multilingual setting \citep{yilmaz2021multi}, but to the best of our knowledge, language-specific class imbalance has not been studied in detail.

In this paper, we analyse the effect of class imbalance \footnote{In this paper, we refer to the non-uniform joint distribution of language and label as "imbalance", even though in the traditional sense imbalance mainly refers to the marginal distribution of labels.} on the model with a number of  experiments of multilingual classification on two different datasets. We chose to work with balanced dataset which we artificially imbalance to allow for controlled experiments. More specifically, we create two subsets of the data, one with a uniform joint distribution of language and labels and one with a skewed one. We want to create a better understanding of the influence of imbalance in multilingual fine-tuning. We first show that imbalance has a negative influence on performance and leads to the latent space becoming more separated by language. Then, using SHAP values, we show evidence that the model learns to encode the imbalance even in non-informative tokens, thus effectively learning to classify based on language identity to an extent. We modify the traditional class weighing method to weigh datapoints of different languages separately and show that this mitigates the negative effects of the imbalance.

In summary, our main contributions are:
\begin{itemize}
    \item We show the detrimental effects of language-specific class imbalance, namely worse performance and a greater separation of languages in the latent space.
    \item Using SHAP values, we show that the model pays more attention to uninformative features when fine-tuned on a dataset with this imbalance, in effect acting more like a language identifier.
    \item We provide a simple method for mitigation by adapting the traditional class weighing method to multilingual fine-tuning.
\end{itemize}

\section{Methods}
\subsection{Text classification}
We use a large language model followed by a classifier head to perform the text classification. For each dataset, we create two subsets of the same size to be used for fine-tuning. One of them, which we will refer to as \textbf{"imbalanced"}, is sampled in a way such that the joint distribution of language and labels is skewed, but the marginal distributions of language and of labels are uniform. The other subset is referred to as \textbf{"balanced"} because the joint distribution of labels and languages is uniform. We sample these subsets such as to maximize the overlap of datapoints between the two to control for the quality of the training data. The test sets for both tasks are balanced.
The classifier head is one feed-forward layer followed by a SoftMax layer. 

\subsection{Language identification}
To analyze the language-specificity of the latent space of the models, we train a logistic regression classifier on the task of identifying the language of text from an external dataset. We use the last CLS token as feature vector. We use sklearn's default parameters, and we report 5-fold cross-validation scores. This is meant to measure how identifiable the languages are in the latent space. We use 1000 articles per language, and we only include the languages the model has seen during fine-tuning. We also report the language identification accuracy on the test sets of the dataset used for fine-tuning.

\subsection{Cumulative difference in SHAP values}
We use SHAP values \citep{NIPS2017_7062} to investigate how the model makes predictions and how this changes between the balanced and imbalanced cases. SHAP values estimate the marginal contributions of each input token by iteratively masking them and observing the changes in the predicted probability. For a given datapoint $\left\{T,y\right\}$ where $T$ is a sequence of tokens $\{t_i\}_{i=0}^{|T|-1}$ and $y$ is a class label, a fine-tuned LLM attributes probability $p(T,y)$ to the event "$T$ belongs to class $y$". SHAP values $S(t)$ explain the contribution of each token $t$ to that probabilitiy according to: 
\begin{equation}
    p(T,y) = \sum_{t\in T}S(t) + b
\end{equation}
$b$ is the value that the model attributes to $p({T_{mask}},y)$ where $T_{mask} = \{{mask}\}_{i=0}^{|T|-1}$, i.e. the probability of label $y$ that the model gives to an input of mask tokens of the same length as T. We name $S_{bal}(t)$ and $S_{imbal}(t)$ the SHAP values calculated from the models trained on the balanced and imbalanced datasets respectively.
We create three subsets of the tokens:
\begin{align*}
    T_{pos} &= \left\{ t_i \in T |S_{bal}(t_i) > 0.01 \right\} \\
    T_{neg} &= \left\{t_i \in T |S_{bal}(t_i) < -0.01 \right\} \\
    T_{neutral} &= \left\{t_i \in T |-0.01 \leq S_{bal}(t_i) \leq 0.01 \right\}
\end{align*}
We calculate the cumulative difference in SHAP value for each set $T_{pos}, T_{neg}$ and $T_{neutral}$ as $\sum_{t \in T} S_{imbal}(t) - S_{bal}(t)$. We calculate this metric for each datapoint in the test set, group them by language and average them.

\subsection{Per-language class weighing}
The traditional class weighing method to address label imbalance in machine learning consists in weighing under- and over-represented labels in the loss such that they count more or less in the gradient calculation\footnote{We attempted other methods for mitigation, namely entropy maximisation and gradient reversal of a language identification head. These methods prevented the model from learning.}. We modify it by applying different class weights for each language and label pair. Let $n_l$ be the number of samples in language $l$, $n_{c,l}$ the number of samples in language $l$ with label $c$, $C$ the total number of classes and $w_{c,l}$ the weight applied to a sample of class $c$ and of language $l$. The weights are calculated according to:
\begin{equation}
    w_{c,l} = \frac{n_l}{C\cdot n_{c,l}}
\end{equation}

\section{Experimental setup}
The language model that we use is Multilingual BERT \citep{devlin-etal-2019-bert}. Specifically, we use the "bert-base-multilingual-cased" model from Huggingface. 
We use a batch size of 16 with a gradient accumulation step of 8. We select the best model according to the validation loss. We use a linearly decreasing learning rate starting at $5e-5$ for the language model and $5e-4$ for the classifier head. They both reach 0 at the end of training\footnote{We make our code available \href{https://github.com/idiap/class-imbalance-multilingual-ft}{here}.}.
We also perform the same experiments with XLM-R \cite{conneau_unsupervised_2020} and report the results in the annex \ref{annex:xlmr}.

We use the Amazon reviews dataset \citep{marc_reviews} in French, German, Spanish, English, Japanese and Chinese, as those are all the available languages in the dataset, and XNLI \citep{conneau-etal-2018-xnli} in French and English, since we wanted to test a bilingual setup. XNLI is a text entailment task. For the Amazon dataset, we train our models to predict the number of stars given (from 1 to 5). For the language identification experiments, we use the Wiki dataset \citep{wikidump}, specifically the pre-processed Wikipedia dataset found on huggingface. The distribution of labels per language for the imbalanced datasets can be seen in Tables \ref{tab:dist_label_amz} and \ref{tab:dist_label_xnli}. The test sets for both XNLI and the Amazon reviews dataset are balanced in both marginal and joint distributions of language and labels.

\begin{table}
\begin{subtable}[h]{\linewidth}
  \begingroup
  \setlength{\tabcolsep}{0.8\tabcolsep}
  \footnotesize
    \begin{tabular}{|c|c|c|c|c|c|}
        \hline
        Star review & 1 & 2 & 3 & 4 & 5 \\
        \hline
        FR,ES,JA & 6.6\% & 13.3\% & 20.0\% & 26.7\% & 33.3\% \\
        DE,EN,ZH & 33.3\% & 26.7\% & 20.0\% & 13.3\% & 6.6\%\\
        \hline
    \end{tabular}
    \endgroup
    \caption{Amazon reviews}
    \label{tab:dist_label_amz}
\end{subtable}

\begin{subtable}[h]{\linewidth}
    \centering
    \begingroup
    \setlength{\tabcolsep}{0.7\tabcolsep}
    \begin{tabular}{|c|c|c|c|c|c|}
        \hline
        Category & Entailment & Neutral & Contradiction \\
        \hline
        FR & 50\% & 33.33\% & 16.67\%  \\
        EN & 16.67\% & 33.33\% & 50\%\\
        \hline
    \end{tabular}
    \endgroup
    \caption{XNLI}
    \label{tab:dist_label_xnli}
\end{subtable}
\caption{Distribution of training and validation set labels for the imbalanced subset.}
\end{table}

\begin{table}[h]
\begin{subtable}[h]{\linewidth}
  \begingroup
  \setlength{\tabcolsep}{0.8\tabcolsep}
  \footnotesize
    \begin{tabular}{|c|c|c|c|c|c|}
        \hline
        Star review & 1 & 2 & 3 & 4 & 5 \\
        \hline
        FR,ES,JA & 13.9\% & 20.6\% & 20.1\% & 19.6\% & 25.8\% \\
        DE,EN,ZH & 27.3\% & 20.7\% & 16.9\% & 20.4\% & 14.7\%\\
        \hline
    \end{tabular}
    \endgroup
    \caption{Amazon reviews}
    \label{tab:dist_label_amz_imbal}
\end{subtable}

\begin{subtable}[h]{\linewidth}
    \centering
    \begingroup
    \setlength{\tabcolsep}{0.7\tabcolsep}
    \begin{tabular}{|c|c|c|c|c|c|}
        \hline
        Category & Entailment & Neutral & Contradiction \\
        \hline
        FR & 32.8\% & 39.0\% & 28.1\%  \\
        EN & 23.1\% & 35.8\% & 41.1\%\\
        \hline
    \end{tabular}
    \endgroup
    \caption{XNLI}
    \label{tab:dist_label_xnli_imbal}
\end{subtable}
\caption{Distribution of test set predictions for the model trained on the imbalanced subset.}
\end{table}

\begin{table}
    \centering
    \begin{tabular}{|c|c|c|}
        \hline
        Training setup & XNLI & Amz. rev. \\
        \hline
        Balanced  & 0.810 & 0.580 \\
        Imbalanced & 0.783 & 0.556 \\
        Imbal. + CW & 0.795 & 0.569 \\
        \hline
       \end{tabular}
    \caption{Test set accuracy for mBERT}
    \label{tab:test_acc_mbert}
\end{table}

\begin{table}
    \centering
    \resizebox{0.48\textwidth}{!}{\begin{tabular}{|c|c|c|c|}
        \hline
        \text{Dataset} & \text{Training setup} & Original & Wikipedia \\
        \hline
        \multirow{3}{*}{Amazon}& Balanced & 0.613 & 0.480 \\
        
         & Imbalanced & 0.847 & 0.646 \\
         & Imbal.+CW & 0.709 & 0.569 \\
         \hline
         \multirow{3}{*}{XNLI}& Balanced & 0.615 & 0.614 \\
         & Imbalanced & 0.928 & 0.899 \\
         & Imbal.+CW & 0.585 & 0.679 \\
        \hline
       \end{tabular}}
    \caption{Language identification average accuracy for mBERT}
    \label{tab:lid_acc_bert}
\end{table}

\begin{figure*}[t!]
\centering
    \begin{subfigure}{0.48\textwidth}
        \centering
        \includegraphics[width=1\linewidth, trim={0 0 70px 0}]{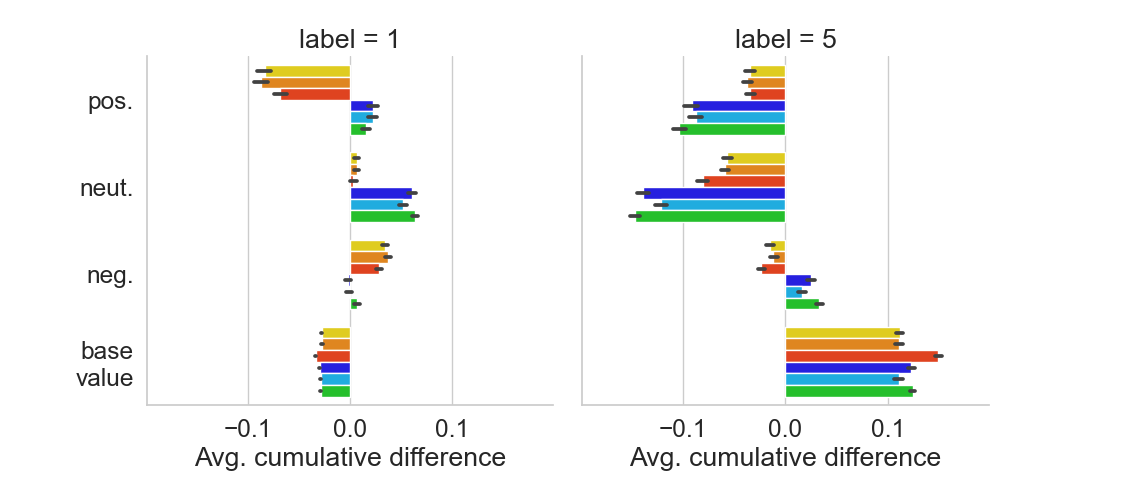}
        \caption{Amazon reviews, Imbalanced}
        \label{fig:amz_bert_imbal}
    \end{subfigure}%
    \begin{subfigure}{0.52\textwidth}
        \centering
        \includegraphics[width=1\linewidth]{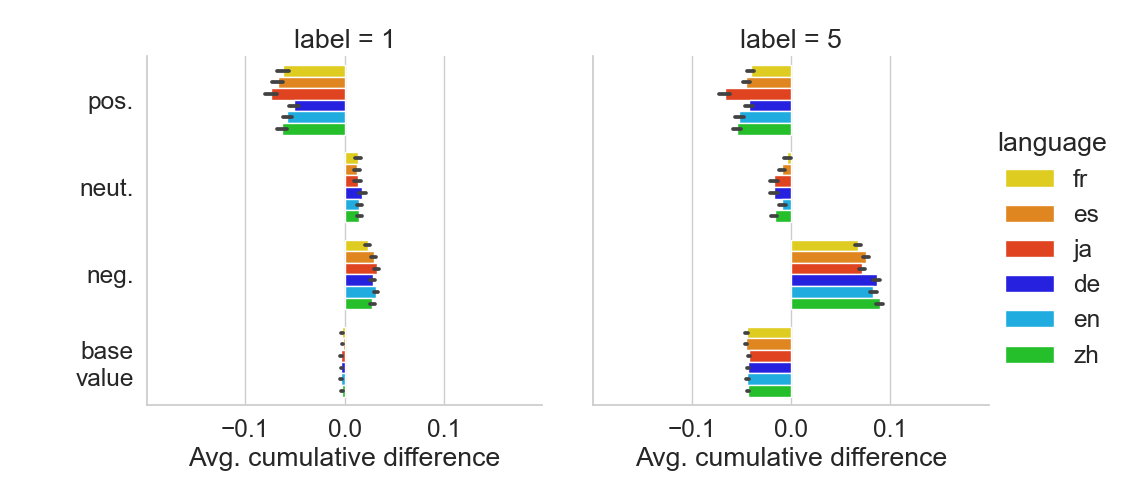}
        \caption{Amazon reviews, Imbal. + CW}
        \label{fig:amz_bert_imbal_cw}
    \end{subfigure}
    \begin{subfigure}{0.48\textwidth}
        \centering
        \includegraphics[width=1\linewidth, trim={0 0 70px 0}]{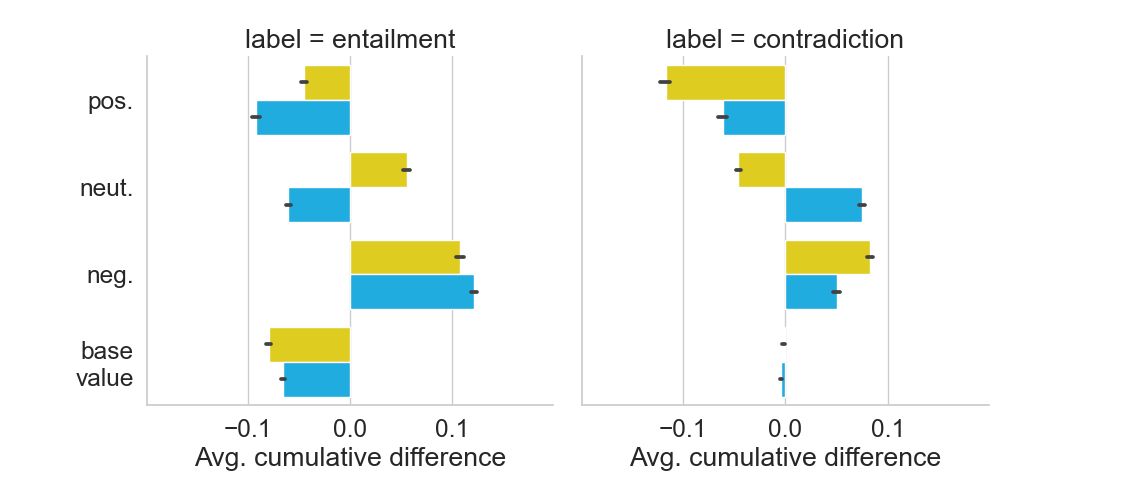}
        \caption{XNLI, Imbalanced}
        \label{fig:xnli_bert_imbal}
    \end{subfigure}%
    \begin{subfigure}[b]{0.52\textwidth}
        \centering
        \includegraphics[width=1\textwidth]{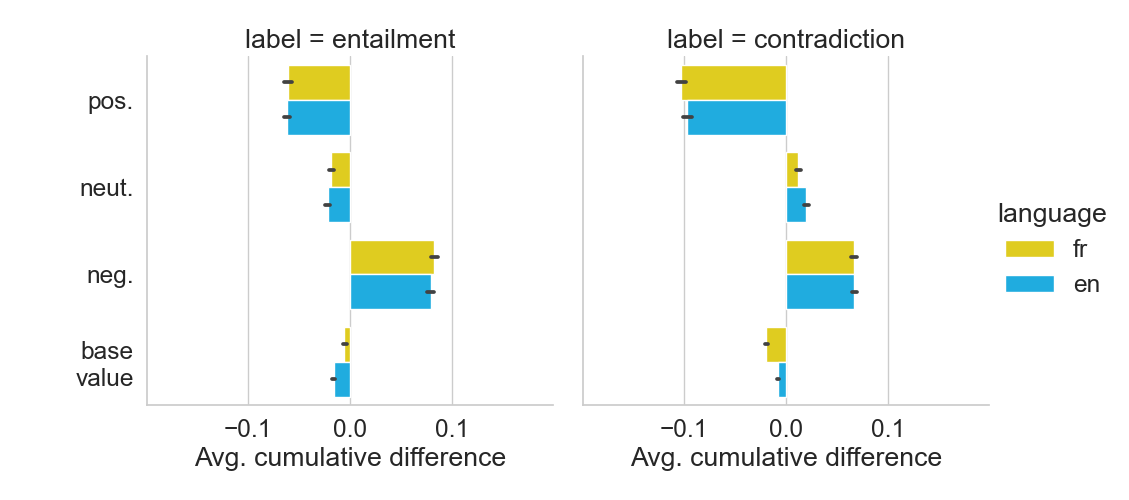}
        \caption{XNLI, Imbal. + CW}
        \label{fig:xnli_bert_imbal_cw}
    \end{subfigure}
    \caption{Average cumulative difference in SHAP value by token category for mBERT.}
    \label{fig:avg_cum_diff_shap}
\end{figure*}

\section{Results and discussion}
In this section, we discuss results with respect to task performance first, after which we will shed light on the effect on language specificity of the multilingual space by showing results from experiments on language identification. We will show in detail what happens to different sets of features when confronted with class imbalance using SHAP values. Lastly, we show that per-language class weighing mitigates the effects of the imbalance.

\subsection{The imbalance worsens performance}
In Table \ref{tab:test_acc_mbert}, we report the test set accuracy for models trained on the balanced and imbalanced datasets.
Unsurprisingly, we see that the models trained on the balanced datasets perform better than the ones trained on the imbalanced datasets.
To understand how the imbalance causes the model to perform worse, we check the distribution of predicted classes by the imbalanced model on the test set in Table \ref{tab:dist_label_amz_imbal}. We see that English, German, and Chinese texts are more likely to have lower reviews, whereas French, Spanish, and Japanese texts are more likely to have higher reviews, thereby following the class distribution in the imbalanced datasets. This seems to indicate that the model learns to make predictions based on language.
In Table \ref{tab:dist_label_xnli_imbal}, we see the same effect with the XNLI labels: English is more likely to be labeled as contradiction, whereas French is more likely to be labeled as entailment.

\subsection{The languages are more identifiable in the latent space}
Ideally, the aim of multilingual fine-tuning is to allow the model to discover patterns across different languages that help it do the task well in a given language. However, if the model learns to rely on language identification rather than patterns that generalize across languages, we expect the latent space to have clearer separation of the languages.
In Table \ref{tab:lid_acc_bert}, we can see that for both XNLI and the Amazon reviews dataset, the language identification accuracy is higher for the model trained on the imbalanced dataset compared with the balanced one. This is further evidence that the model focuses more on the language of the input in the presence of imbalance.

\subsection{The model learns to rely on non-informative tokens}
Knowing that the languages become more distinct in the latent space in the presence of imbalance, we want to use SHAP values to analyze how the model makes predictions at the token level.

\subsubsection{Amazon reviews}
In Figure \ref{fig:amz_bert_imbal}, on the left side, the average cumulative difference in SHAP value for label 1 of the Amazon reviews dataset is shown. French, Spanish and Japanese positive tokens contribute more negatively in the imbalanced case, and negative tokens contribute more positively. Thus, tokens that had a high absolute SHAP value in the balanced case now have a lower absolute value in the imbalanced case for these under-represented languages. The model relies less on features that were informative in the balanced case for these languages.
For the over-represented languages, the main effect is that neutral tokens now contribute positively to the prediction. The model thus sees non-informative tokens in the over-represented language as an indication of that label.

On the right side, we see the same plot for label 5. There is a significant difference in base value which we attribute to model artifacts. This means that the SHAP values in the imbalanced case will have a negative bias since the base value is much higher for that model. Thus, the difference in SHAP value between that model and the balanced one will also have a negative bias\footnote{We discuss this issue further in Annex \ref{annex:shap_bias} and introduce a way to mitigate it.}. However, we can still see that the under- and over-represented language groups are treated differently: positive and neutral tokens for the over-represented languages become less negative than for the under-represented ones, and neutral token become more positive for the over-represented and more negative for the under-represented.

\subsubsection{XNLI}
Figure \ref{fig:xnli_bert_imbal} shows the same plots for the XNLI dataset. French is over-represented for the "entailment" label, and English is over-represented for the "contradiction" label. For both labels, the neutral tokens contribute more positively for the over-represented language and more negatively for the under-represented. The model again learns to rely on non-informative tokens from the over-represented language. For the under-represented language, the positive tokens contribute more negatively, and the negative ones more positively. Their absolute SHAP values are thus lower and the model again learns to rely less on informative tokens for this language. It is actually also the case for the over-represented language but to a lesser extent. This simply points to the fact that the model is paying less attention to informative features overall and more attention to the language of the input.
\\

The overall trend in both XNLI and the Amazon reviews dataset is that positive tokens contribute more negatively and negative token contribute more positively. Neutral tokens contribute either positively if they are of the over-represented languages or negatively if they are of the under-represented languages. Thus, the model puts less importance on features that were relevant in the balanced case and treats the simple presence of non-informative tokens of a certain language as indication of a certain label, in effect acting more like a language identifier. 

\subsection{Per-language class weighing mitigates the effect of the imbalance}

First, Table \ref{tab:test_acc_mbert} shows that overall performance on the tasks improves with class weighing on imbalanced data. Also, in Table \ref{tab:lid_acc_bert}, we see that language identification scores are lower with the class weighing method than without, being almost on-par with the balanced case. 

Figure \ref{fig:amz_bert_imbal_cw} and \ref{fig:xnli_bert_imbal_cw} show that while the cumulative difference in SHAP value is not null, it is on average smaller than without the weighing. We still see that positive tokens are less positive, and negative tokens are more positive, i.e. SHAP values of relevant features are smaller in this case than in the balanced case, but we do not see a clear separation between over- and under-represented languages like we do in the imbalanced case. Moreover, the difference in SHAP values for neutral tokens is minimal, which means that uninformative tokens stay irrelevant for the model. 

Overall, we see that the per-language class weighing method mitigates the effects of the language-specific class imbalance: the latent space is less separated by language and the model does not learn to treat tokens from under- and over-represented languages differently.

\section{Conclusion}

In this paper, we showed that a language model trained on a seemingly balanced multilingual dataset, with uniform marginal distributions of languages and of labels, but skewed joint distribution of language and label, will learn this skew. We first showed that the model performs worse in the presence of this imbalance. Based on the distribution of the test set predictions, we show that it learns to make predictions based on language, which can negatively impact its out-of-distribution performance. We also showed that the imbalance leads to the latent space being more separated by language. We then analyzed SHAP values to better understand how the way the model makes predictions changes. SHAP values showed that features that the model used when trained on balanced data became less important when trained on imbalanced data, and that features that were "neutral", i.e. didn't contribute to the prediction of a given label, became more important.
We modify the traditional method of class weighing by calculating class weights separately for each language and train a model on the imbalanced dataset with a weighted loss. We show that this simple method is effective at mitigating the negative effects of the imbalance. 

This is of high stakes, as training on multiple languages is often done in real-life cases, and preventing the perpetuation of biases is often desirable. It is a reminder that large language models and deep learning architectures in general do not necessarily follow human intuition and will make predictions based on what is available in the data. 
Training a model to build robust features requires careful consideration of not just the marginal distribution of the dataset features but also of their joint distribution. 

\section{Limitations}

A main limitation of our study is the artificial nature of our datasets. These datasets have equal representation of languages and labels, which allowed us to isolate the issue of language-label imbalance. However, in real-life datasets, one will often face imbalances both in the marignal and joint distributions.

Another limitation is the sole use of SHAP values for our explainability method. We used Layer Integrated Gradients but we would not be able to show cumulative values which show an overall picture of the effects. However, according to \cite{atanasova_diagnostic_2020}, occlusion methods like SHAP are only worse than gradient-based methods in terms of their computational efficiency. 

Our method for per-language class weighing simply modifies the traditional class weighing method. However, as seen in \cite{henning-etal-2023-survey}, newer weighing method exist which could also have been adapted and led to improved performance. 

\newpage
\bibliography{anthology, custom, references}
\bibliographystyle{acl_natbib}

\appendix

\section{Appendix}

\subsection{Dataset statistics}

\begin{table}[ht]
    \centering
    \begin{tabular}{|c|c|c|}
        \hline
        Split & XNLI & Amz. rev. \\
        \hline
        Train & 524k & 719k \\
        \hline
        Validation & 3.3k & 18k \\
        \hline
        Test & 6.6k & 30k \\
        \hline
       \end{tabular}
    \caption{Number of datapoints for train, validation and test split}
    \label{tab:number_datapoints}
\end{table}

\subsection{Results on XLM-Roberta}
\label{annex:xlmr}
We report the results from the same experiments performed with mBERT, with XLM-R. Test set accuracy is shown in Table \ref{tab:test_acc_xlmr}, language identification accuracy is shown in Table \ref{tab:lid_acc_xlmr} and cumulative difference in SHAP values is shown in Figure \ref{fig:avg_cum_diff_shap_xlmr}.
Across the board, we can see that the findings from the mBERT results also apply to XLM-R: imbalance makes the latent space more distinct, it promotes uninformative features and demotes relevant ones, and per-language class weighing can help mitigate those effects.
The XLM-R models have been trained with an added loss to minimize their difference in base value to make the results more interpretable, which is explained in \ref{annex:shap_bias}.

\begin{table}
    \centering
    \begin{tabular}{|c|c|c|}
        \hline
        Training setup & XNLI & Amz. rev. \\
        \hline
        Balanced  & 0.829 & 0.596 \\
        Imbalanced & 0.812 & 0.586 \\
        Imbalanced + CW & 0.828 & 0.594 \\
        \hline
       \end{tabular}
    \caption{Test set accuracy for XLM-R}
    \label{tab:test_acc_xlmr}
\end{table}

\begin{table}
    \centering
    \resizebox{0.48\textwidth}{!}{\begin{tabular}{|c|c|c|c|}
        \hline
        \text{Dataset} & \text{Training setup} & \text{Original} & \text{Wikipedia} \\
        \hline
        \multirow{3}{*}{Amazon}& Balanced & 0.309 & 0.389 \\
         & Imbalanced & 0.381 & 0.744 \\
         & Imbal.+CW & 0.412 & 0.503 \\
         \hline
         \multirow{3}{*}{XNLI}& Balanced & 0.556 & 0.582 \\
         & Imbalanced & 0.838 & 0.865 \\
         & Imbal.+CW & 0.605 & 0.607 \\
        \hline
       \end{tabular}}
    \caption{Language identification average accuracy for XLM-R}
    \label{tab:lid_acc_xlmr}
\end{table}

\subsection{SHAP value bias due to difference in base values}
\label{annex:shap_bias}
\begin{figure*}[h]
\centering
    \begin{subfigure}{0.48\textwidth}
        \centering
        \includegraphics[width=1\linewidth, trim={0 0 70px 0}]{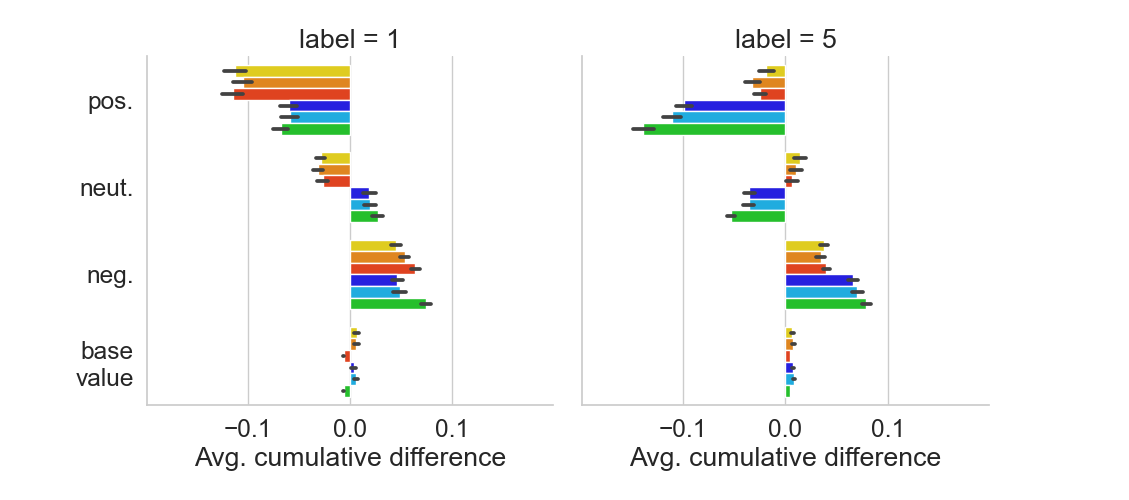}
        \caption{Amazon reviews, Imbalanced}
        \label{fig:amz_xlmr_imbal}
    \end{subfigure}%
    \begin{subfigure}{0.52\textwidth}
        \centering
        \includegraphics[width=1\linewidth]{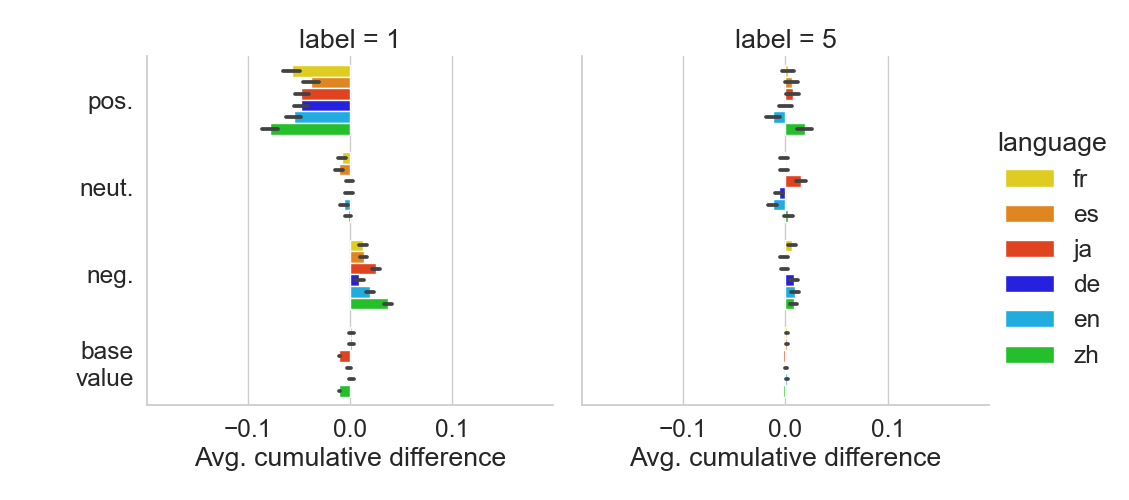}
        \caption{Amazon reviews, Imbal. + CW}
        \label{fig:amz_xlmr_imbal_cw}
    \end{subfigure}
    \begin{subfigure}{0.48\textwidth}
        \centering
        \includegraphics[width=1\linewidth, trim={0 0 70px 0}]{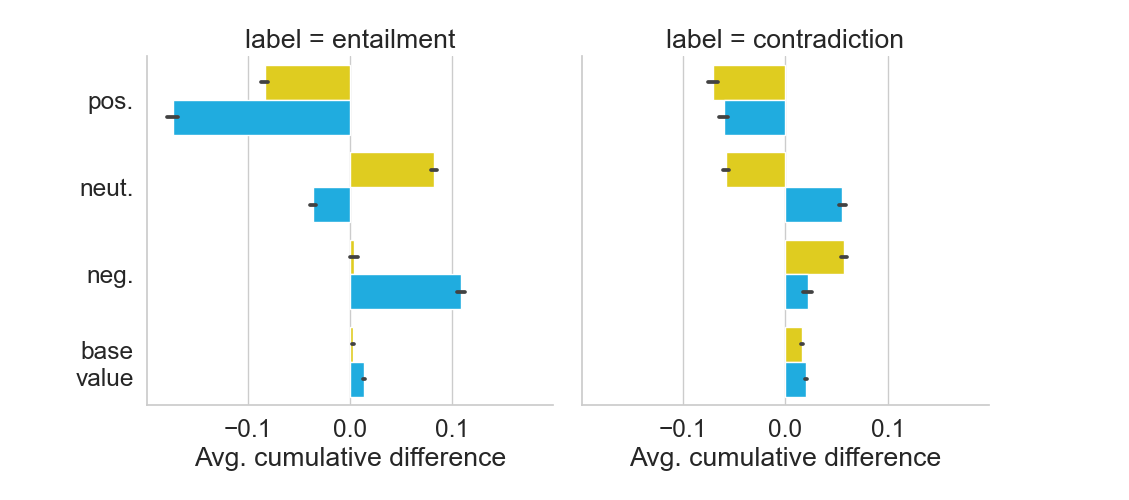}
        \caption{XNLI, Imbalanced}
        \label{fig:xnli_xlmr_imbal}
    \end{subfigure}%
    \begin{subfigure}[b]{0.52\textwidth}
        \centering
        \includegraphics[width=1\textwidth]{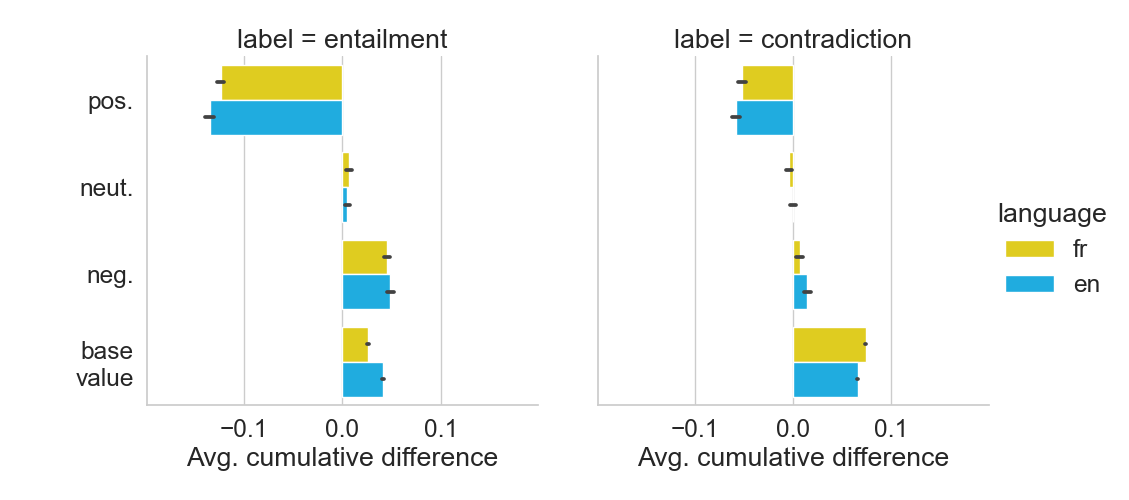}
        \caption{XNLI, Imbal. + CW}        \label{fig:xnli_xlmr_imbal_cw}
    \end{subfigure}
    \caption{Average cumulative difference in SHAP value by token category for XLM-R with the added masked input entropy maximisation loss.}
    \label{fig:avg_cum_diff_shap_xlmr}
\end{figure*}

\begin{figure*}[h]
\centering
    \begin{subfigure}{0.48\textwidth}
        \centering
        \includegraphics[width=1\linewidth, trim={0 0 70px 0}]{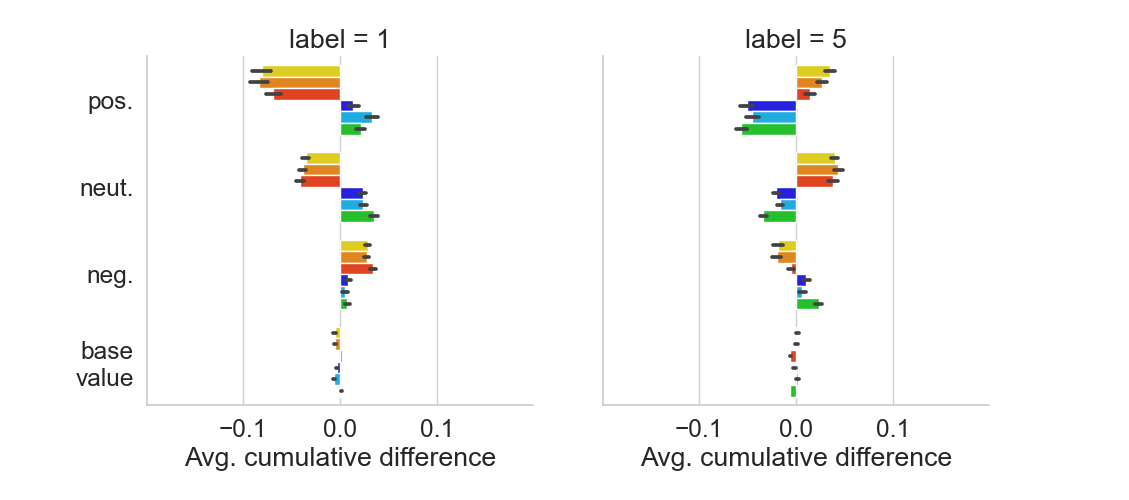}
        \caption{Amazon reviews, Imbalanced}
        \label{fig:amz_bert_imbal_mask_ent}
    \end{subfigure}%
    \begin{subfigure}[b]{0.52\textwidth}
        \centering
        \includegraphics[width=1\linewidth]{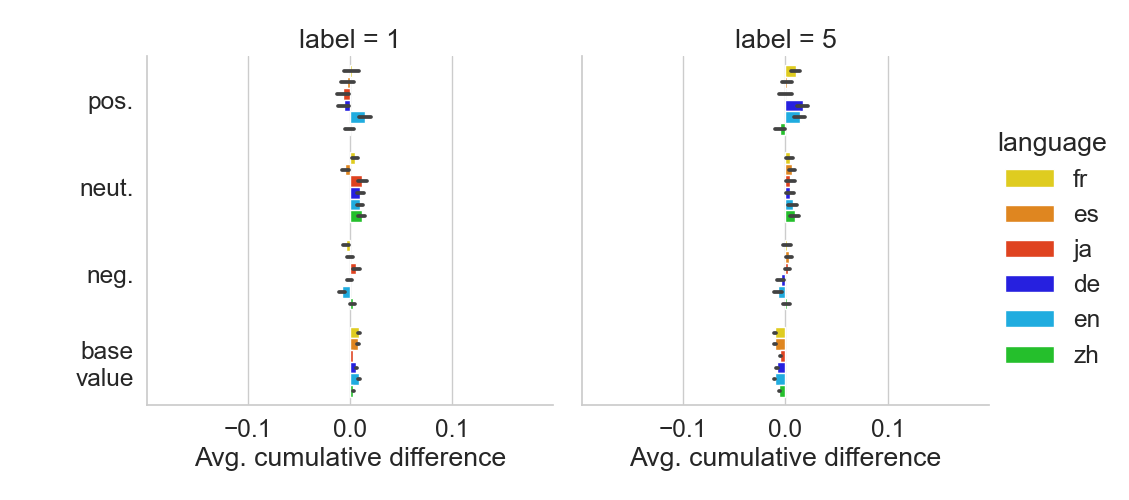}
        \caption{Amazon reviews, Imbal. + CW}
        \label{fig:amz_bert_imbal_cw_mask_ent}
    \end{subfigure}
    \begin{subfigure}{0.48\textwidth}
        \centering
        \includegraphics[width=1\linewidth, trim={0 0 70px 0}]{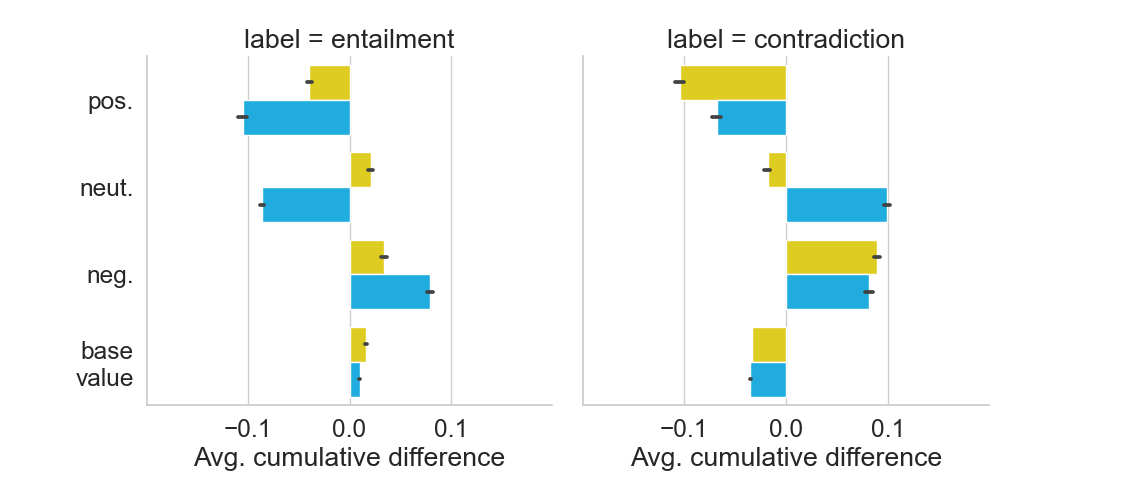}
        \caption{XNLI, Imbalanced}
        \label{fig:xnli_bert_imbal_mask_ent}
    \end{subfigure}%
    \begin{subfigure}[b]{0.52\textwidth}
        \centering
        \includegraphics[width=1\linewidth, trim={0 0 0 0}]{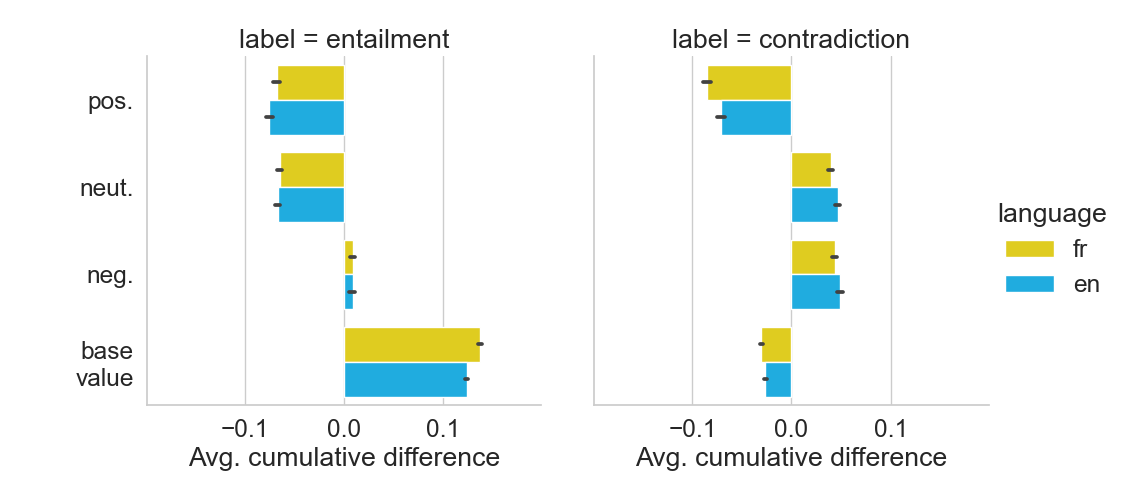}
        \caption{XNLI, Imbal. + CW}
        \label{fig:xnli_bert_imbal_cw_mask_ent}
    \end{subfigure}
    \caption{Average cumulative difference in SHAP value by token category for mBERT with the added masked input entropy maximisation loss.}
    \label{fig:avg_cum_diff_shap_mbert_maskent}
\end{figure*}

One of the main issues we faced using SHAP values is that they are not easily comparable across models due to the difference in base values. In the current implementation of SHAP values, the base values are calculated by replacing every token in the input by the mask token and taking the output probabilities. Ideally, we want those probabilities to be the same across models. 
To achieve this, we added the entropy of the output distribution of a fully masked input multiplied by -1 to the loss at every gradient step, so as to incentivise the model to output a uniform distribution. Let $M(T): \mathbb{R}^{L\times d} \rightarrow \mathbb{R}^C$ be the model we use for prediction, where $T$ is the input of token embeddings of length $L$, $d$ is the dimension of the embedding and $C$ is the number of classes and $\mathbb{C}$ the set of classes. Let $m$ be an input of mask tokens of length between 1 and $L$. We add the following loss to the total loss:
\begin{equation}
     l = \sum_{c\in \mathbb{C}} M(m)[c] \cdot \log(M(m)[c])
\end{equation}
We refer to it as the masked input entropy loss.
We find that this does not hinder downstream task performance, but makes differences in base values much smaller, making the cumulative difference in SHAP values much easier to interpret. We show the same plots as in Figure \ref{fig:avg_cum_diff_shap} with models trained with this added loss in Figure \ref{fig:avg_cum_diff_shap_mbert_maskent}. We also only show the XLM-R results with this added loss in Figure \ref{fig:avg_cum_diff_shap_xlmr}.

\subsection{Justification for threshold}

We set our threshold at a SHAP value of 0.01 for what we consider neutral and positive/negative tokens as this resulted in an approximate 20/60/20 (neg./neut./pos.) split. We experimented with a threshold of 0.001 and 0.05. The first one did not include enough tokens in the neutral token groups for the cumulative difference in SHAP value to make sense. The second one showed similar results in the cumulative difference in SHAP values, just with slightly different magnitudes. Our analysis most likely still holds with higher thresholds, up to a point. We had considered regression-type analysis between the SHAP values of models trained on balanced and imbalanced data because they would not have required the addition of a threshold. However, they would not have allowed us to capture the cumulative effect of the change in SHAP values.

\end{document}